\documentclass[times,twocolumn,final]{elsarticle}
\usepackage{framed,multirow}

\usepackage{amssymb}
\usepackage{latexsym}
\usepackage{amsmath}

\usepackage{url}
\usepackage{xcolor}

\usepackage{hyperref}

\definecolor{newcolor}{rgb}{.8,.349,.1}

\begin{document}

%\verso{Chengliang Dai \textit{et~al.}}

\begin{frontmatter}

\title{Suggestive Annotation of Brain MR Images with Gradient-guided Sampling}%
%%\tnotetext[tnote1]{This is an example for title footnote coding.}

\author[1]{Chengliang Dai\corref{cor1}}
\cortext[cor1]{Corresponding author: Chengliang Dai
e-mail:c.dai@imperial.ac.uk}
\author[1]{Shuo Wang}
%%\fntext[fn1]{This is author footnote for second author.}
\author[1]{Yuanhan Mo}
%% Third author's email
%%\ead{author3@author.com}
\author[2]{Elsa Angelini}
\author[1]{Yike Guo}
\author[1,3]{Wenjia Bai}

\address[1]{Data Science Institute, Imperial College London, United Kingdom}
\address[2]{NIHR Imperial Biomedical Research Centre, Imperial College London, United Kingdom}
\address[3]{Department of Brain Sciences, Imperial College London, United Kingdom}

%%\received{1 May 2021}
%%\finalform{10 May 2013}
%%\accepted{13 May 2013}
%%\availableonline{15 May 2013}
%%\communicated{S. Sarkar}

\begin{abstract}
%%%
Machine learning has been widely adopted for medical image analysis in recent years given its promising performance in image segmentation and classification tasks. The success of machine learning, in particular supervised learning, depends on the availability of manually annotated datasets. For medical imaging applications, such annotated datasets are not easy to acquire, it takes a substantial amount of time and resource to curate an annotated medical image set. In this paper, we propose an efficient annotation framework for brain MR images that can suggest informative sample images for human experts to annotate. We evaluate the framework on two different brain image analysis tasks, namely brain tumour segmentation and whole brain segmentation. Experiments show that for brain tumour segmentation task on the BraTS 2019 dataset, training a segmentation model with only 7\% suggestively annotated image samples can achieve a performance comparable to that of training on the full dataset. For whole brain segmentation on the MALC dataset, training with 42\% suggestively annotated image samples can achieve a comparable performance to training on the full dataset. The proposed framework demonstrates a promising way to save manual annotation cost and improve data efficiency in medical imaging applications.  
%%%%
\end{abstract}


\end{frontmatter}

%\linenumbers

\section{Introduction}
\label{S:1}
Medical imaging plays an significant role in many clinical applications such as: disease diagnosis, clinical decision making, treatment planning and patient management. There are hundreds of millions of medical imaging scans conducted every year around the world creating a huge amount of available data. In England, there were 43.5 million imaging tests reported in the 12 months from March 2018 to March 2019 \cite[]{england2016diagnostic}. The increasing number of medical images inevitably makes the number of experienced radiologists to interpret these complex images fall short of demand. The recent development of deep learning techniques can help to alleviate the shortage of qualified radiologists. Extensive research has shown that deep learning techniques can achieve performance comparable to or even surpassing human-level performance in a variety of tasks \cite[]{liang2019evaluation}, particularly in medical imaging applications such as segmentation \cite[]{liu2020deep}, registration \cite[]{haskins2020deep} and reconstruction \cite[]{knoll2020deep}.

Despite the growing popularity of deep learning, there are several major challenges; one of them being the scarcity of data required for training deep learning models. It is well-known that the good performance of deep learning models often relies on the large amount of data used for training. The real challenge in medical imaging applications is not just the availability of imaging data itself, but also the scarcity of expert annotations for these images. Medical image annotation can be expensive in terms of both time and required expertise. For example, annotating a 3D brain tumour image for a single patient may take several hours even for a skilled image analyst \cite[]{fiez2000lesion}.

Several methods have shown promising results in alleviating the data scarcity challenge in medical imaging by training deep learning models from limited annotations. For instance, unsupervised and semi-supervised learning methods aim to leverage information from abundant unannotated samples and to improve the accuracy and robustness of the model using these samples \cite[]{cheplygina2019not}. Generative adversarial network (GAN) based approaches \cite[]{zhao2019data,yi2019generative,ravanbakhsh2020human} are widely adopted for generating synthetic images and pseudo annotations; images that can be used for model training. The accuracy and robustness of the models are improved after training synthetic datasets compared to training on the original dataset alone \cite[]{zhao2019data,yi2019generative,ravanbakhsh2020human}. Some methods such as dataset distillation \cite[]{wang2018dataset} aim to reduce the size of training dataset by creating very few synthetic samples that carry as much information as the original large dataset. Moreover, transfer learning and domain adaptation methods have been proposed to learn features from one domain where a large amount of annotated data is available and then transfer these features to a new domain where annotated data is limited \cite[]{tajbakhsh2016convolutional,kamnitsas2017unsupervised}.

\begin{figure*}
\centering
\includegraphics[width=0.99\textwidth]{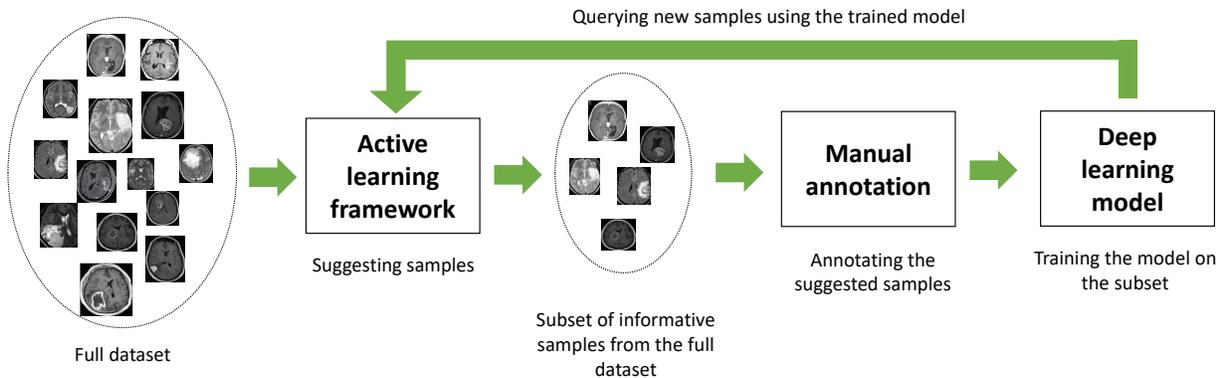}
\caption{A typical framework of active learning. A full dataset without annotation is fed to an active learning framework, which suggests a subset of the unannotated dataset to be annotated by the experts. The annotated subset can then be used to train a deep learning model. The suggestion can be done iteratively until the deep learning model reaches a satisfactory performance for the downstream task.} \label{activelearning}
\end{figure*}

Most methods mentioned previously are developed for harvesting as many features or samples as possible from existing datasets to enhance the model performance without acquiring more annotations. On the other hand, an active learning (AL) approach tackles the problem by suggesting an optimal subset of an unannotated dataset that is effective and beneficial for model training. A typical AL framework is demonstrated in Fig. \ref{activelearning}.

In this paper, we propose a gradient-guided sampling mechanism that can be used directly by a general AL framework to utilize the gradient of the loss term from the deep learning model for suggesting informative samples for manual annotation. We explored two brain image segmentation tasks, brain tumour segmentation and whole brain segmentation. In both tasks, our method is capable of suggesting samples that are more helpful to training a segmentation model compared to state-of-the-art methods, which allow the model to produce comparable performance to a model trained on the full dataset.

\subsection{Related work}
Selecting informative data samples is the essence of any AL framework. There are many approaches used in the area of AL for quantifying the informativeness of a data sample from the underlying data distribution for a given task. Depending on how informativeness is evaluated, these approaches can be mostly categorised into two groups: sampling by uncertainty and sampling by representativeness \cite[]{yang2015multi,yang2017suggestive,budd2019survey}.

\subsubsection{Uncertainty Sampling}
The simplest and most intuitive way is to suggest samples about which it is uncertain how to annotate. By asking human experts to annotate these samples and including these annotations in the training set, more information can be learnt by the model. This is known as the uncertainty method \cite[]{settles2009active}. The uncertainty of a sample can be quantified in different ways; one way is to quantify uncertainty by the class probability of its most likely label. Samples whose most likely label have the lowest probability are most uncertain, these are referred to as the least confident method in \cite[]{settles2009active}. For image segmentation tasks, this method can be extended to measuring the sum of lowest class probability for each pixel in a given image segmentation \cite[]{budd2019survey}. More certain predictions will have higher pixel-wise class probabilities. Therefore, a lower sum of the maximal class probability over all pixels in an image would indicate a higher uncertainty of prediction. However, the least confident method suffers the problem of under-representing the distribution of less likely labels, as only information about the most likely labels are considered. Marginal sampling was proposed as an improved version of the least confident sampling method \cite[]{settles2009active}, this considers the first and second most probable labels for multi-class scenarios and calculates the difference between them. Another commonly adopted approach is to use the entropy of the probability over all labels for quantifying the uncertainty \cite[]{settles2009active}.

Ensemble-based uncertainty measurement \cite[]{yang2017suggestive,jungo2019assessing} is another direction for uncertainty estimation, it requires training multiple models or applying test-time data augmentation such as Monte-Carlo (MC) dropout for estimation. Ensemble-based methods usually measure the agreement between different models that perform the same task; assuming data samples that cause more disagreement between models imply a higher level of uncertainty and informativeness.

Another option for measuring uncertainty is by predicting the training loss \cite[]{yoo2019learning}. The method adds a separate task head to the backbone network, which is trained for predicting the training loss of the downstream task. This allows the training loss to be estimated on the unannotated data and data samples with higher predicted loss can be sampled for annotation \cite[]{yoo2019learning}. Furthermore, Bayesian networks have been explored to model the uncertainty of predictions either directly  \cite[]{gal2017deep} or by incorporating geometric smoothness priors \cite[]{konyushkova2019geometry}.

\subsubsection{Representativeness Sampling}
Representativeness sampling strategies are mostly used in addition to uncertainty-based methods \cite[]{yang2017suggestive,yang2015multi,budd2019survey} to avoid sampling from the same area of the data distribution and creating a biased model towards this particular area.

\cite[]{yang2017suggestive} proposed to sample uncertain data samples that better represent the full dataset by solving a maximum set-cover (max-cover) problem on image feature vectors. In \cite[]{zheng2019biomedical}, the authors adopted a hierarchical clustering method in addition to the max-cover method, in which sampling was performed in a latent space learnt from generative models such as the generative adversarial network (GAN) \cite[]{goodfellow2014generative} or the variational autoencoder (VAE) \cite[]{kingma2013auto}. In \cite[]{zheng2019biomedical}, uncertainty is not measured in order to decouple the sampling and segmentation processes. In \cite[]{smailagic2018medal}, the distance between feature descriptors are calculated to make sure that the sampled images are as dissimilar to each other as possible so that a large area of the data distribution can be covered. Similarly in \cite[]{zhou2017fine}, a patch-level diversity score is calculated in order to select a sample that is different from the previous one.

\subsection{Contributions}
Inspired by the success of adversarial learning \cite[]{goodfellow2014explaining,kurakin2016adversarial} using adversarial samples to improve the efficiency of model training, we propose a gradient-guided sampling method to suggest informative data samples for manual annotation on a data manifold learnt by the generative model. 

Major contributions of this work include:
\begin{itemize}
\item We incorporate the loss gradient into the suggestive annotation process, an area that has not been investigated in depth for active learning in medical image segmentation tasks.
\item We utilise a generative model, the VAE, to model the data distribution on a latent space. This allows the loss gradient to be projected from a high-dimensional image space into a low-dimensional latent space so sampling can be performed more efficiently.
\item The proposed method suggests a batch of unannotated data samples based on the training loss of the samples suggested in the previous iteration and does not require ranking for the unannotated data samples based on their levels of uncertainty or representativeness.
\item We re-implement and evaluate some of the state-of-the-art active learning methods published previously for different tasks on brain image segmentation task.
\end{itemize}

A preliminary version of this work was presented at MICCAI 2020 \cite[]{dai2020suggestive}. Here we have substantially extended the conference paper, including:
\begin{itemize}
    \item Previously, we only investigated the scenario of suggesting one batch of samples for manual annotation. To explore the full potential of the proposed framework, here we extend to an iterative sampling and annotation scenario.
    \item Previously, the framework was tested on a single-class brain tumour segmentation task on the BraTS dataset. Here we extend to the more challenging multi-class segmentation task. 
    \item Apart from brain tumour segmentation task, we further include a new dataset and evaluate the method performance for whole brain segmentation task.
    \item Furthermore, we add other previously published state-of-the-art methods for comparison.
\end{itemize}

\section{Methods}
In our framework, we assume that there is a large pool of unannotated medical images \(U=\{x_1,x_2,...,x_n\}\) and there is one or multiple experts from whom we can request manual annotation for a data sample $x_{U} \in U$. A labelled training set \(L\) normally contains all the data samples from \(U\) and their annotations. The proposed framework aims to find an optimal subset \(L' \subseteq L\) so that a model trained on \(L'\) can achieve comparable performance to the model trained on \(L\).

\begin{figure*}
\centering
\includegraphics[width=0.99\textwidth]{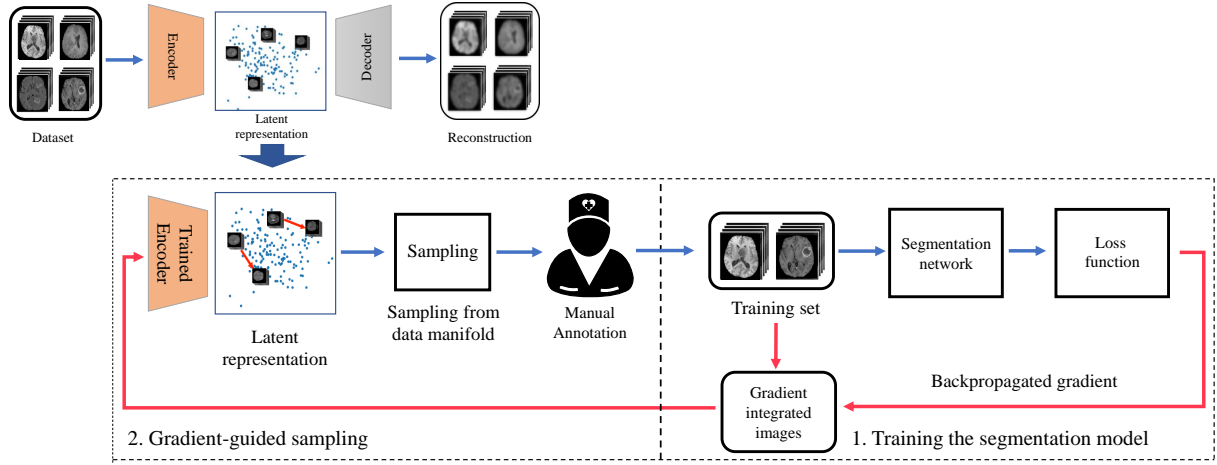}
\caption{An overview of the gradient-guided suggestive annotation framework. The top panel illustrates the learning of a data manifold for the full dataset using a VAE, which enables the projection of samples from image space to latent space. The bottom panel illustrates the iterative steps of the framework: (1) The segmentation network is trained using suggested samples. (2) Explore the latent space to suggest samples for annotation, guided by the gradient of loss from the segmentation network. The loss gradient is integrated with samples in the image space and then projected to the latent space for exploration.} \label{framework}
\end{figure*}

An overview of the proposed framework is illustrated in Fig. \ref{framework}. To utilise the gradient of loss from the segmentation model for sampling, we first train a VAE on the unannotated dataset \(U\) to learn a data manifold. The encoder of the VAE enables the projection of image samples onto the manifold. Then the sampling process consists of two steps: (1) Train an image segmentation model \(f(x|L')\) in an active learning manner, where \(L'\) denotes the set of annotated samples suggested by the framework. (2) Suggest unannotated samples by exploring the latent space. After training for several epochs, the gradient of the segmentation training loss is backpropagated to the image space, then projected to the latent space using the VAE encoder. Unannotated samples are explored in the latent space and suggested based on the loss gradient. They are referred to experts for annotation and added to \(L'\). The framework iterates between Step (1) and Step (2), gradually incrementing the labelled set for training the model. 

Step (1) and (2) are performed iteratively to increment the labelled subset \(L'\) that contains the most informative data samples given a intermediate model \(f(x|L')\) and an unannotated dataset \(U\). Both the model and the annotated dataset \(L'\) will evolve over time within this framework.

\subsection{Learning the Data Manifold}
A data manifold that reflects the structure of the unannotated dataset is fundamental to the proposed framework. In this work, the VAE is used for learning the data manifold given its good potential shown in other works \cite[]{liu2017unsupervised}. The VAE is trained on \(U\) with the loss function formulated as
\begin{equation}\label{eq:vaeloss}
\begin{aligned}
\mathcal{L}_{vae}(\theta,\phi;x_{i}) ={} & MSE(f_{\theta}(g_{\phi}(x_{i})),x_{i}) \\
& + D_{KL}(q_\phi(z|x_{i})||p_\theta(z)),
\end{aligned}
\end{equation}

\noindent where \(g_{\phi}(\cdot)\) and \(f_{\theta}(\cdot)\) denote the encoder and decoder, which typically consist of a number of convolutional layers \cite[]{kingma2013auto}. MSE denotes the mean square error function and \(D_{KL}\) denotes the KL-divergence, which regularises the optimisation problem by minimising the distance between the latent variable distribution and a Gaussian distribution \cite[]{kingma2013auto}. Once trained, the VAE can be used to obtain the latent representations of samples \(Z=\{z_1,z_2,...,z_n\}\) given \(U\), which will be used for the main framework explained below.

\subsection{Step (1): Training the Segmentation Model}
A segmentation model will be trained, using the labelled set $L'$. For the first iteration, a set of $m$ random samples will be annotated and used to initialise the model training. For all subsequent iterations, informative samples will suggested, annotated and added to $L'$ to re-train the model. This is consistent with the observations in \cite[]{kumar2010self, bengio2009curriculum} that less \textit{informative} samples are more important at the early stage of model training, while \textit{harder} samples are more important at the later stage. 

An annotated dataset containing m initial samples \(L'=\{(x_{1},y_{1}),(x_{2},y_{2}),...,(x_{m},y_{m})\}\) is thus constructed, where $y$ denotes the annotation by the expert. Unlike the single Dice loss function we used in \cite[]{dai2020suggestive}, the segmentation model is trained on \(L'\) using the multi-class Dice loss function defined by,
\begin{equation}\label{eq:dice}
\mathcal{L}_{Dice}(y_{i},\hat{y}_{i})= -\frac{2}{|K|}\sum_{k\in K}\frac{\hat{y}_{ik}y_{ik}}{\hat{y}_{ik}^{2}+y_{ik}^{2}},  (x_{i},y_{i})\in L'.
\end{equation}
where \(\hat{y}_{i}\) denotes the segmentation probability map for image \(x_{i}\) and \(y_{i}\) denotes the ground truth annotation by the expert. \(\hat{y}_{i}\) and \(y_{i}\) contain K channels of probabilities for K classes. \(\hat y_{ik}\) denotes the probability map for class \(k \in K\) and \(y_{ik}\) denotes the probability for class k from the one-hot encoding of ground truth.

\subsection{Step (2): Gradient-guided Sampling}
After the segmentaion model is trained, for each of the $m$ samples \(x_{i}\in L' \), its gradient of loss is backpropagated to the image space according to,
\begin{equation}\label{eq:integration}
    x'_{i}=x_{i}+\alpha \cdot \frac{\partial L_{Dice}}{\partial x_{i}},
\end{equation}
\noindent where the gradient of the loss function informs the direction to more informative samples and \(\alpha\) denotes the step length along the gradient. 

Then we sample informative data samples on the data manifold and strikes a balance between exploring uncertain samples and representative samples. Using the VAE encoder \(g_{\phi}(\cdot)\), the hard sample \(x'_{i}\) can be projected to the latent space by,
\begin{equation}\label{eq:projection}
    z'_{i}=g_{\phi}(x'_{i}).
\end{equation}

\begin{figure}[!t]
\centering
\includegraphics[scale=.24]{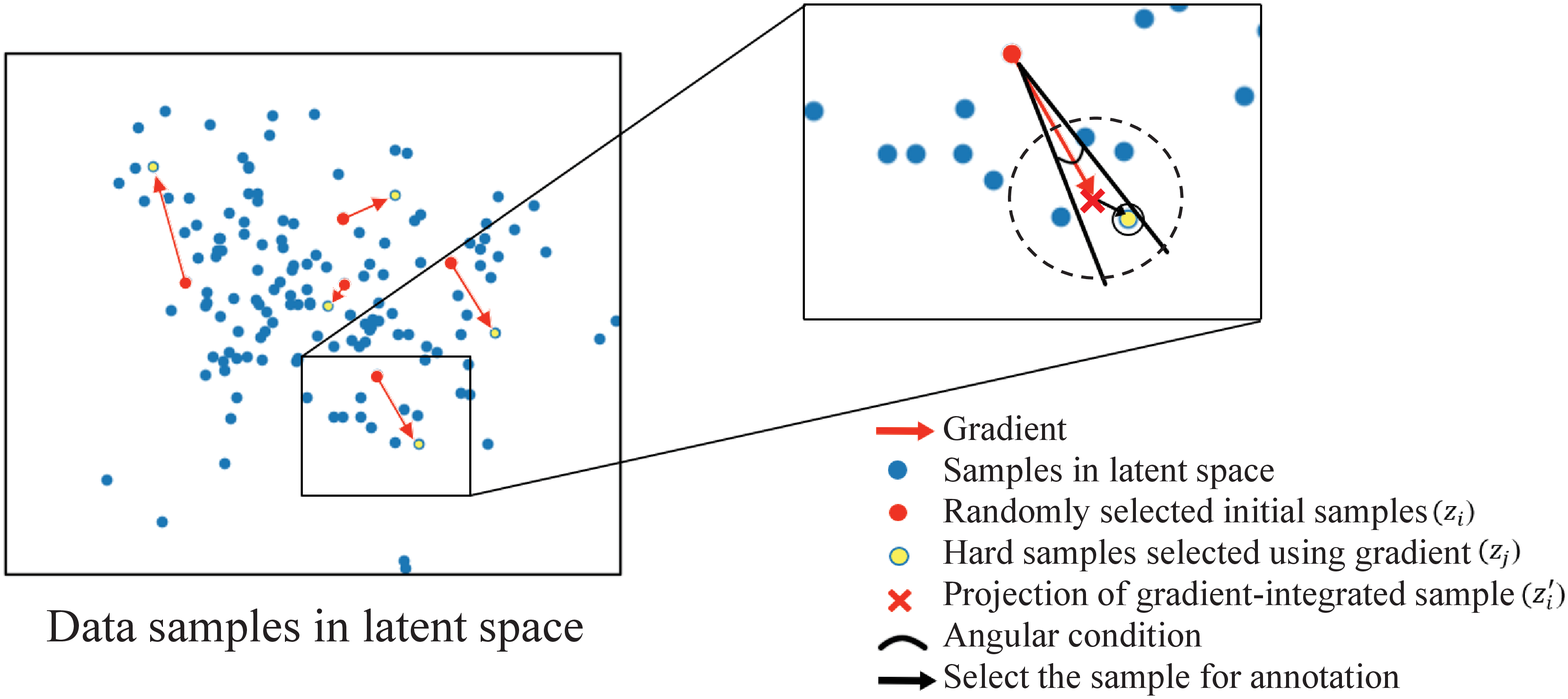}
\caption{Sampling process in the latent space. The zoomed view shows an example of a newly suggested sample. The red dot denotes a random initial sample \(z_{i}\). The red cross denotes the gradient-integrated sample \(z'_{i}\). The proposed method searches along the gradient direction (red arrow) within a region defined by the angular condition and suggests the new sample \(z_{j}\) (yellow dot) for manual annotation.} \label{latent}
\end{figure}

Fig. \ref{latent} illustrates the sampling process in the latent space. The blue dots represent samples from the unannotated dataset \(U\). The red dots are the latent representation of the samples suggested in the previous iteration. After training the segmentation model with these samples in the previous iteration, for any \(x_{i}\) from \(L'\), the loss gradient is backpropagated to the image space and integrated to \(x_{i}\) (Eq. \ref{eq:integration}), then projected to \(z'_{i}\) in the latent space (Eq. \ref{eq:projection} and red cross in Fig. \ref{latent}). We empirically define two criteria for suggestion. First, an existing real image should be close to \(z'_{i}\) in the latent space, which is to find a \(z_{j}\in Z\) (yellow dot) that has the shortest Euclidean distance to \(z'_{i}\) using
\begin{equation}\label{eq:euclidean}
    d(z_{j},z_{i}')=|z_{j}-z_{i}'|.
\end{equation} Second, an angular condition is introduced to limit the search angle (black angle in Fig. \ref{latent}). If the selected sample does not satisfy the angular condition, we check the sample that has the second shortest Euclidean distance. The angular condition is defined using the cosine distance widely used in machine learning:
\begin{equation}\label{eq:angular}
    cos(\theta)=\frac{(z_{j}-z_{i}')\cdot(z_{i}'-z_{i})}{||z_{j}-z_{i}'||\cdot||z_{i}'-z_{i}||}.
\end{equation}
The angular condition was used to make sure the search would not deviate from the gradient direction. 

Using the gradient-guided sampling method, one or more samples can be found for each \(z'_{i}\). We only select one sample \(z_{j}\in Z\) in the neighbourhood of \(z'_{i}\) in the latent space, which corresponds to image \(x_{j}\in X\) in the image space. In this way, we sample $m$ informative data samples in the first iteration and suggest them to the expert for manual annotation. An updated training dataset with $m$ more samples \(L'=\{(x_{1},y_{1}),(x_{2},y_{2}),...,(x_{2m},y_{2m})\}\) can be constructed for second iteration, which consists of $m$ initial samples and $m$ new samples suggested by the proposed method. The enlarged training set can be used for fine-tuning the segmentation model. If we repeat the training and suggesting steps iteratively for $n$ times, the training set would contain \((n+1)*m\) data samples in the end.

Theoretically, images can be synthesised from $z'_{i}$ via the VAE decoder \(f_{\theta}(\cdot)\) and suggested to the expert for annotation. However, the synthetic image may not be of high quality in reality, which would prevent the expert from producing reliable annotation. To mitigate this issue, here we sample in the real image space, searching for existing real images that are most similar to the synthesised image in the latent space. In this way, the expert would be able to annotate on high-quality real images.

\section{Experiments}
\subsection{Datasets}
We evaluated the proposed method on two public brain MRI datasets.

\noindent
\textit{\textbf{A. Brain Tumour Segmentation Challenge (BraTS) 2019}}: This dataset contains T1, T1 gadolinium (Gd)-enhanced, T2 and T2-FLAIR brain MRI volumes of 335 patients diagnosed with high-grade gliomas (HGG) or low-grade gliomas (LGG), acquired with different clinical protocols from multiple institutions \cite[]{bakas2017segmentation}. The dataset was pre-processed with skull-striping, interpolation to a uniform isotropic resolution of 1$\times$1$\times$1 \(mm^{3}\) and registered to the SRI24 template space \cite[]{rohlfing2010sri24} with a dimension of $240\times240\times155$.  During pre-processing, the first and last few image slices (normally blank) were discarded the resulting dimension of the image is $256\times256\times150$. The images were manually annotated with four labels: background, the peritumoural edema, the Gd-enhanced tumour (ET) and the necrotic and non-enhancing tumour core. The latter two labels were combined to form the tumour core (TC) label and latter three labels were combined to form the whole tumour (WT) label, which were the structures for evaluating segmentation accuracy in the BraTS challenge \cite[]{bakas2018identifying}. The dataset was randomly split into 260/75 for training and test, respectively.

\noindent
\textit{\textbf{B. Multi-Atlas Labelling Challenge (MALC)}}: The dataset contains brain MRI scans from 30 subjects with manual annotations of the whole brain for 134 fine-grained regions including the background \cite[]{landman2012miccai}. A coarse-grained segmentation map that contains 25 major brain structures (110 sub-regions are combined to create structure cortical gray matter left and right) was generated and evaluated in this work. The 25 major brain structures were previous chosen by multiple researchers \cite[]{wachinger2018deepnat,roy2019quicknat} for investigating the performance of the segmentation algorithms. The dataset was split into 20/10 for training and test, respectively. 

We further processed both BraTS and MALC datasets with zero padding and z-score intensity normalisation.

\subsection{Experimental Design}
For MR image segmentation tasks, the framework can be applied to either 3D or 2D networks. Due to the memory limit of standard graphics processing units (GPUs) that was available to us (NVIDIA Tesla P100, 16GB RAM), we utilised 2D networks in this work for image segmentation. 

A VAE with the similar architecture from \cite[]{liu2017unsupervised} was used for learning the data manifold. The VAE has four levels of blocks and each level contains one residual block with batch normalization and leaky ReLU (with a leakiness of 1e-2) activation function. It was trained for 50 epochs on both datasets with the Adam optimizer, a learning rate of 1e-4, a mini-batch size of 16, and the loss function defined in Eq. \ref{eq:vaeloss}. A large $z$-dimension would result in a sparsely populated latent space, making gradient-guided suggestion difficult. Therefore, we empirically set the $z$-dimension to 3 for the BraTS dataset and 2 for the MALC dataset as the MALC dataset is much smaller than the BraTS dataset. 

The vanilla U-Net architecture from \cite[]{ronneberger2015u} was chosen for image segmentation with minor changes when training on the brain imaging dataset. The U-Net has four levels of blocks. Each block contains two convolutional layers with batch normalization and ReLU activation function, one max pooling layer in the encoding part and de-convolutional layers in the decoding part. The number of convolutional filters in each block is: 32, 64, 128, and 256; and the bottleneck layer has 512 convolutional filters. The U-Net was trained with the Adam optimizer using the loss function given by Eq. \ref{eq:dice}, a learning rate of 1e-3 and a mini-batch size of 32. 

Data augmentation including flipping, random intensity and random elastic deformation were conducted on-the-fly during training. The step length $\alpha$ in Eq. \ref{eq:dice} was set to 1e-3 empirically.

To initialise the segmentation network for the BraTS dataset, we randomly sampled 0.5\% data samples from the BraTS dataset to train the network for 10 epochs. Then we applied the proposed sampling method to sample 0.5\% more data samples from the dataset for manual annotation and added the annotated samples to the training set, making it 1\% of the full dataset. The segmentation network was trained on the enlarged training set for 10 more epochs. We performed suggestive sampling and network training for 8 more iterations, each time adding 1\% data samples with manual annotations to the training set, followed by network training. 

For the experiment on the MALC dataset, the segmentation network for the MALC dataset was initialised and trained in a similar fashion. Due to the relatively small size of the MALC dataset, each iteration we added 6\% data samples suggested by the method to the training set and performed network training for 10 epoch. We repeated the same training and suggestion processes for 8 more iterations on MALC datasets, adding 6\% more data samples to the MALC training dataset each time.

We compared five different suggestive annotation methods: random suggestion, the deep active learning framework (DAL) \cite[]{yang2017suggestive}, the clustering and max-cover based method (ClsMC) \cite[]{zheng2019biomedical}, the learning loss method (LL) \cite[]{yoo2019learning} and the oracle method. Random suggestion is the baseline method, which randomly selects data samples for training the segmentation model each time. DAL suggests both uncertain and representative samples. In each iteration, uncertain samples are selected using bootstrapping. Representative samples are selected by suggesting samples that bear useful features for as many unannotated samples as possible based on the cosine distance between high level features of candidate samples. \cite[]{yang2017suggestive}. ClsMC suggests samples based on the representativeness qualified by adopting hierarchical clustering and max-cover algorithms in the latent space. The suggestion is made one-time only \cite[]{zheng2019biomedical}. LL method selects samples that have the highest training loss predicted by a network using U-Net as the backbone and a separate head for predicting the training loss. For the LL method \cite[]{yoo2019learning}, samples are suggested iteratively in the same fashion as the proposed method. DAL, ClsMC and LL methods were originally proposed for different applications or on different datasets from this work. Where the original code was not available, we re-implemented DAL, ClsMC and LL methods and adopted the same experiment setting to enable a fair comparison of method performance.

For the DAL re-implementation, we made some modifications to cope with the experimental design in this paper. We replaced the FCN in the original framework with the same U-Net used by our framework. We trained 4 U-Nets to measure the uncertainty score. The High level features were extracted from the bottleneck layers of the U-Nets for measuring the cosine similarity. Due to the large number of samples suggested in each iteration, we had to empirically choose the number of candidate samples and perform max-cover algorithm multiple times in each iteration to reduce the computational complexity (e.g., if 200 samples are needed in an iteration, we select 20 samples with the highest uncertainty scores from the unannotated dataset and perform max-cover algorithm to suggest 10 samples from 20 candidate samples. The suggestion process is done 20 times in order to suggest 200 samples from 400 candidate samples.). We trained the U-Nets following the same procedure of the proposed method. Instead of selecting a best model using validation set, the result of the ensemble of 4 U-Nets was used for evaluation.

For the ClsMC re-implementation, we followed the steps described in \cite[]{zheng2019biomedical} by first conducting agglomerative clustering on the latent representations, for deciding how many clusters we would select images from. We then selected images from clusters using max-cover strategy and further select most representative images from previously selected images. We set the size of the candidate set in each cluster following a similar approach described in \cite[]{zheng2019biomedical} taking the computational complexity into account. The pixel ratio defined in \cite[]{zheng2019biomedical} is replaced with the percentage of full dataset used. The suggestion process stops when the size of the training set reaches a certain percentage of the full dataset. The U-Net trained on the samples selected by ClsMC and the training configuration are the same as the proposed method.

For the LL re-implementation, we attached a similar network structure for the loss prediction to the U-Net backbone and used the same loss prediction method as described in \cite[]{yoo2019learning}, which calculates the difference between a pair of loss predictions. The loss function for segmentation network is Dice loss. The margin and scaling constant for the loss function defined in \cite[]{yoo2019learning} were set to 0.5 and 1 respectively. The whole LL network was trained for 50 epochs using the same learning rate, optimiser and batch size as the U-Net model used by other methods. We stopped propagating the gradient from the loss prediction module to the segmentation module after 30 epochs following a similar procedure mentioned in \cite[]{yoo2019learning}.

The oracle method serves as an empirical upper-bound for suggestive annotation. It assumes that the annotations are already known for all samples, and we choose the posteriori `best' samples according to the downstream task. For segmentation tasks, we use the most challenging samples for the current segmentation model with lowest predicted Dice scores. To allow a fair comparison, each experiment was repeated 10 times and the average performance was evaluated. 

\section{Results}
\subsection{Parameter settings}
An optimal manifold representation of the dataset and a sampling mechanism that can maximise the effectiveness of using gradient of loss from the segmentation network are the two most essential elements in the proposed framework. Consequently, we experimented with several parameter settings for learning the data manifold and for sampling. For the data manifold, we tested different dimensions for the latent space $z$ in training the VAE. For the sampling mechanism, we explored sampling new data samples on the data manifold using Euclidean distance only and with an additional angular condition. The results of using different parameter settings are reported below.

\subsubsection{Dimensionality of the Latent Space}
We found that a higher $z$-dimension number brings a better quality to the image reconstruction of the VAE, but a more challenging data manifold for the subsequent sampling process. The results of using different dimensions of $z$ are compared in Table~\ref{tab:zdim}. It shows that the performance of using a 3-dimensional latent space outperforms using a 10-dimensional and a 50-dimensional spaces. A 50-dimensional space leads to an exceptionally good image reconstruction quality for VAE, which is potentially caused by an overfitting latent space given the size of the training set. The segmentation result of using such a high dimensional space is close to random suggestion method on the BraTS dataset as shown in Table~\ref{tab:zdim}. The inferior performance of a higher $z$ dimension number is possibly because that in a high-dimensional latent space, a limited amount of data samples will be sparsely distributed and searching samples along the gradient direction in such a space is challenging.

\begin{table}
\centering
\caption{Influence of the latent space $z$ dimension on the method performance. The performance is evaluated using the mean Dice scores of three tumour structures (whole tumour, enhancing tumour and tumour core) when training the segmentation model using samples suggested in the latent space. 3\% or 6\% of the full BraTS dataset are suggested. Figures in bold are the best results among using a $z$-dimension number of 3, 10 and 50.}\label{tab:zdim}
\scalebox{0.7}{\begin{tabular}{ccccccc}
\hline
                 & \multicolumn{3}{c}{3\% dataset}                    & \multicolumn{3}{c}{6\% dataset}                    \\ \hline
$z$ dimension    & 3                                & 10     & 50     & 3                                & 10     & 50     \\ \hline
Whole Tumour     & \textbf{0.8238} & 0.8081 & 0.8013 & \textbf{0.8437} & 0.8259 & 0.8195 \\
Enhancing Tumour & \textbf{0.6972} & 0.6783 & 0.6692 & \textbf{0.7182} & 0.7049 & 0.6898 \\
Tumour Core      & \textbf{0.7619} & 0.7542 & 0.7479 & \textbf{0.7895} & 0.7687 & 0.7582 \\ \hline
\end{tabular}}
\end{table}

\subsubsection{Ablation Study on Angular Condition}
We investigated the effectiveness of angular condition for sampling. Samples on the data manifold are suggested with or without adding the angular condition. The segmentation performance is compared in Table \ref{tab:angular}. For suggesting with angular condition, an empirical threshold of $\pi/3$ is set for the angular condition. It shows that searching samples with angular condition outperforms that without angular condition. Our experiments indicate that sampling without the angular condition is highly likely to cause the direction of searching new samples on the data manifold to deviate from the gradient, undermining the effectiveness of using gradient in the framework. Therefore, angular condition is used throughout the rest of the experiments in this work.

\begin{table*}
\centering
\caption{Influence of the angular condition on the method performance, evaluated using the mean Dice scores for three tumour structures when training the segmentation model using suggested samples. 3\% or 6\% of the full BraTS dataset are suggested. Figures in bold are the results of using random method, our method withut angular condition and our method with angular condition.}\label{tab:angular}

\begin{tabular}{ccccccc}
\hline
                 & \multicolumn{3}{c}{3\% dataset}                                                                                                     & \multicolumn{3}{c}{6\% dataset}                                                                                                    \\ \hline
Methods          & Random & \begin{tabular}[c]{@{}c@{}}W/o angular \\ condition\end{tabular} & \begin{tabular}[c]{@{}c@{}}With angular\\ condition\end{tabular} & Random & \begin{tabular}[c]{@{}c@{}}W/o angular\\ condition\end{tabular} & \begin{tabular}[c]{@{}c@{}}With angular\\ condition\end{tabular} \\ \hline
Whole Tumour     & 0.8124 & 0.8173                                                           & \textbf{0.8238}                                                           & 0.8225 & 0.8243                                                          & \textbf{0.8437}                                                           \\
Enhancing Tumour & 0.6857 & 0.6723                                                           & \textbf{0.6972}                                                           & 0.7062 & 0.6985                                                          & \textbf{0.7182}                                                           \\
Tumour Core      & 0.7509 & 0.7431                                                           & \textbf{0.7619}                                                           & 0.7703 & 0.7562                                                          & \textbf{0.7895}                                                           \\ \hline
\end{tabular}
\end{table*}

\subsection{Brain Tumour Segmentation on BraTS dataset}
\begin{figure}
\centering
\includegraphics[width=0.48\textwidth]{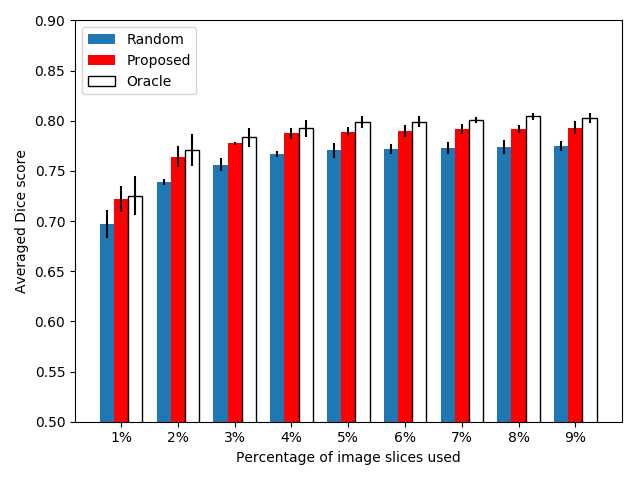}
\caption{Comparison of the proposed method with the random suggestion method and the oracle method. For each iteration, data samples of 1\% of the full BraTS dataset are suggested and added into the training set.} \label{brats_test}
\end{figure}

We first evaluated the proposed framework for brain tumour segmentation task. Fig. \ref{brats_test} compares the performance of proposed method to the random suggestion method and the oracle method (upper bound), in terms of the average Dice score across three tumour structures. It shows that with the increase in the size of labelled dataset, the performance of the segmentation model improves for all three methods and the proposed framework consistently outperforms the random suggestion. The performance increase becomes slower after the training set is expanded to 5\% of the full dataset. The proposed method achieves slightly inferior performance than the oracle method.

We further compared the performance of the proposed method to other suggestive sampling methods, including DAL, ClsMC and LL. In this comparison, all methods suggested 3\% and 7\% of the full dataset. Tables \ref{tab:brats_dice} and \ref{tab:brats_hausdorff} compare the mean Dice score and Hausdorff distance of the methods for each tumour type. We also verified the statistical significance of the results by performing the Wilcoxon signed-rank test between two top performing methods among random, DAL, ClsMC, LL and ours. The result of the Wilcoxon signed-rank tests are included in Tables \ref{tab:brats_dice} and \ref{tab:brats_hausdorff}. It shows that the proposed method yields slightly inferior performance as the oracle method and it outperforms random, ClsMC and LL methods in terms of Dice score. For Hausdorff distance, the proposed method outperforms random, ClsMC and LL methods for segmenting whole tumour and tumour core. Furthermore, by training using only 7\% of the suggested samples, the segmentation performance is close to the performance of training on the full dataset.

The random method outperforms the ClsMC method in some cases as shown in Tables \ref{tab:brats_dice} and \ref{tab:brats_hausdorff}. This is most likely because a random method is quite a strong baseline. As discussed in \cite[]{kumar2010self, bengio2009curriculum}, less informative samples are more important at the early stage of the model training. Nonetheless, ClsMC suggests samples only once based on the pre-defined similarity/coverage scores, therefore it is not feasible for ClsMC to keep improving the quality of suggested samples based on the performance of the segmentation model.

\begin{figure*}
\centering
\includegraphics[width=0.8\textwidth]{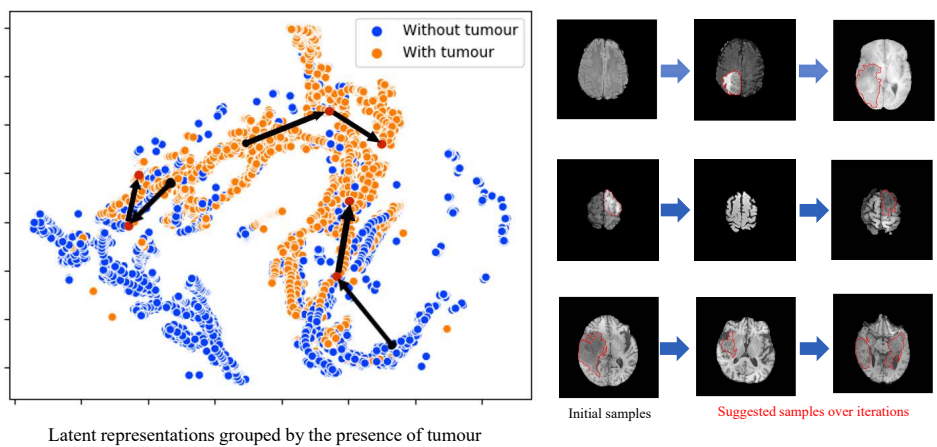}
\caption{Qualitative illustration of the proposed method. The left panel shows the t-SNE plot of the latent space for images from the BraTS dataset. The black arrow denotes the gradient direction, starting from an initial sample (black point) and pointing to a suggested sample (red point). The right panel shows the images corresponding to the starting point and ending point of the black arrow.} \label{tsne}
\end{figure*}

Fig. \ref{tsne} illustrates the t-SNE plot of the latent space for images from the BraTS dataset. The right side of Fig. \ref{tsne} shows three examples of the sampling process, including the initial images that provide the loss gradient and the suggested images after search in the latent space in the following two iterations. As it shows, the suggested images tend to present different intensity contrasts or different anatomical structures from the initial images, which are likely to provide more information for segmentation model training.

\begin{table*}
\centering
\caption{Comparison of the proposed method with other suggestion methods in terms of average Dice score and standard deviation, using 3\% and 7\% of full training set of BraTS dataset. The figures in bold are the best results among random, DAL, ClsMC, LL and our method. The results of training on the full training set and the Wilcoxon signed-rank test between two top performing methods among random, DAL, ClsMC, LL and ours are also given in this table.\label{tab:brats_dice}}
\scalebox{0.8}{
\begin{tabular}{ccccccc}
\hline
\multirow{2}{*}{Methods} & \multicolumn{3}{c}{3\% dataset}                           & \multicolumn{3}{c}{7\% dataset}                           \\ \cline{2-7} 
                         & Whole tumour      & Enhancing tumour  & Tumour core       & Whole tumour      & Enhancing tumour  & Tumour core       \\ \hline
Random                   & 0.8124$\pm$0.0172 & 0.6857$\pm$0.0237 & 0.7509$\pm$0.0189 & 0.8402$\pm$0.0197 & 0.7035$\pm$0.0201 & 0.7701$\pm$0.0177 \\
DAL \cite[]{yang2017suggestive}                      & 0.8156$\pm$0.0146 & 0.6892$\pm$0.0193 & 0.7535$\pm$0.0211 & 0.8457$\pm$0.0133 & 0.7093$\pm$0.0181 & 0.7795$\pm$0.0134 \\
ClsMC \cite[]{zheng2019biomedical}                    & 0.8113$\pm$0.011  & 0.6823$\pm$0.0122 & 0.7587$\pm$0.0131 & 0.8369$\pm$0.0154 & 0.7018$\pm$0.0186 & 0.7801$\pm$0.0164 \\
LL \cite[]{yoo2019learning}                       & 0.8207$\pm$0.0125 & 0.6874$\pm$0.0171 & 0.7562$\pm$0.014  & 0.8491$\pm$0.0106 & 0.7103$\pm$0.0193 & 0.7746$\pm$0.0129 \\
Proposed                 & \textbf{0.8238$\pm$0.0101} & \textbf{0.6972$\pm$0.0103} & \textbf{0.7619$\pm$0.0137} & \textbf{0.8589$\pm$0.0093} & \textbf{0.7291$\pm$0.0127} & \textbf{0.7947$\pm$0.0105} \\
Oracle                   & 0.8297$\pm$0.0141 & 0.7062$\pm$0.0149 & 0.7649$\pm$0.017  & 0.8614$\pm$0.0101 & 0.7396$\pm$0.0116 & 0.7998$\pm$0.0112 \\
Full dataset             & -                 & -                 & -                 & 0.8663$\pm$0.0081 & 0.7457$\pm$0.0104 & 0.8026$\pm$0.0093 \\
p-value                 & $\approx$0.27                  &  $\approx$0.05                 & $\approx$0.04                  &\textless0.01                   &\textless0.01                   &\textless0.01                   \\ \hline
\end{tabular}}
\end{table*}

\begin{table*}
\centering
\caption{Comparison of the proposed method with other methods in terms of averaged Hausdorff distance (unit: mm) and standard deviation, using 3\% and 7\% of full dataset. The figures in bold are the best results among random, DAL, ClsMC, LL and our method. The results of training on the full training set and the Wilcoxon signed-rank test between two top performing methods among random, DAL, ClsMC, LL and ours are also given in this table.\label{tab:brats_hausdorff}}
\scalebox{0.8}{
\begin{tabular}{ccccccc}
\hline
\multirow{2}{*}{Methods} & \multicolumn{3}{c}{3\% dataset}                     & \multicolumn{3}{c}{7\% dataset}                     \\ \cline{2-7} 
                         & Whole tumour    & Enhancing tumour & Tumour core    & Whole tumour   & Enhancing tumour & Tumour core     \\ \hline
Random                   & 12.35$\pm$1.49  & \textbf{16.05$\pm$2.37}  & 20.73$\pm$2.89 & 10.29$\pm$1.26 & \textbf{14.75$\pm$1.79}  & 19.54$\pm$1.87  \\
DAL \cite[]{yang2017suggestive}                      & 12.01$\pm$1.23  & 16.56$\pm$2.1  & 20.03$\pm$2.95 & 10.27$\pm$1.04 & 15.39$\pm$1.8  & 19.23$\pm$2.13  \\
ClsMC \cite[]{zheng2019biomedical}                    & 11.97$\pm$1.27  & 16.43$\pm$2.05   & \textbf{19.81$\pm$2.4} & 10.58$\pm$1.12 & 15.24$\pm$1.92   & 19.06$\pm$1.95  \\
LL \cite[]{yoo2019learning}                       & 12.21$\pm$1.2   & 16.37$\pm$2.17   & 19.97$\pm$2.67 & 10.96$\pm$0.97 & 15.43$\pm$1.67   & 19.12$\pm$2.09  \\
Proposed                 & \textbf{11.42$\pm$1.15} & 16.13$\pm$2.03   & 19.89$\pm$2.53 & \textbf{9.03$\pm$1.18} & 15.04$\pm$1.71   & \textbf{18.51$\pm$1.93} \\
Oracle                   & 11.03$\pm$1.09  & 15.82$\pm$1.9    & 19.42$\pm$2.16 & 9.43$\pm$0.95  & 14.89$\pm$1.32   & 18.11$\pm$1.77  \\
Full dataset             & -               & -                & -              & 8.99$\pm$0.87  & 14.58$\pm$1.19   & 17.54$\pm$1.41  \\
p-value                 &$\approx$0.01                 &$\approx$0.47                  &$\approx$0.59                &$\approx$0.02                &$\approx$0.53                  &$\approx$0.02                 \\ \hline
\end{tabular}}
\end{table*}

\subsection{Whole Brain Segmentation on the MALC dataset}
\begin{figure}
\centering
\includegraphics[width=0.48\textwidth]{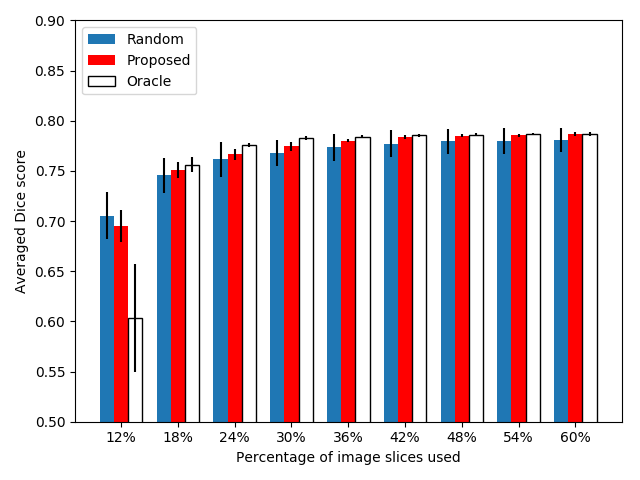}
\caption{Comparison of the proposed suggestive annotation method with the random suggestion method and the oracle method. For each iteration, data samples of 6\% of the full MALC dataset are suggested and added into the training set.} \label{malc_test}
\end{figure}

\begin{figure*}
\centering
\includegraphics[width=0.99\textwidth]{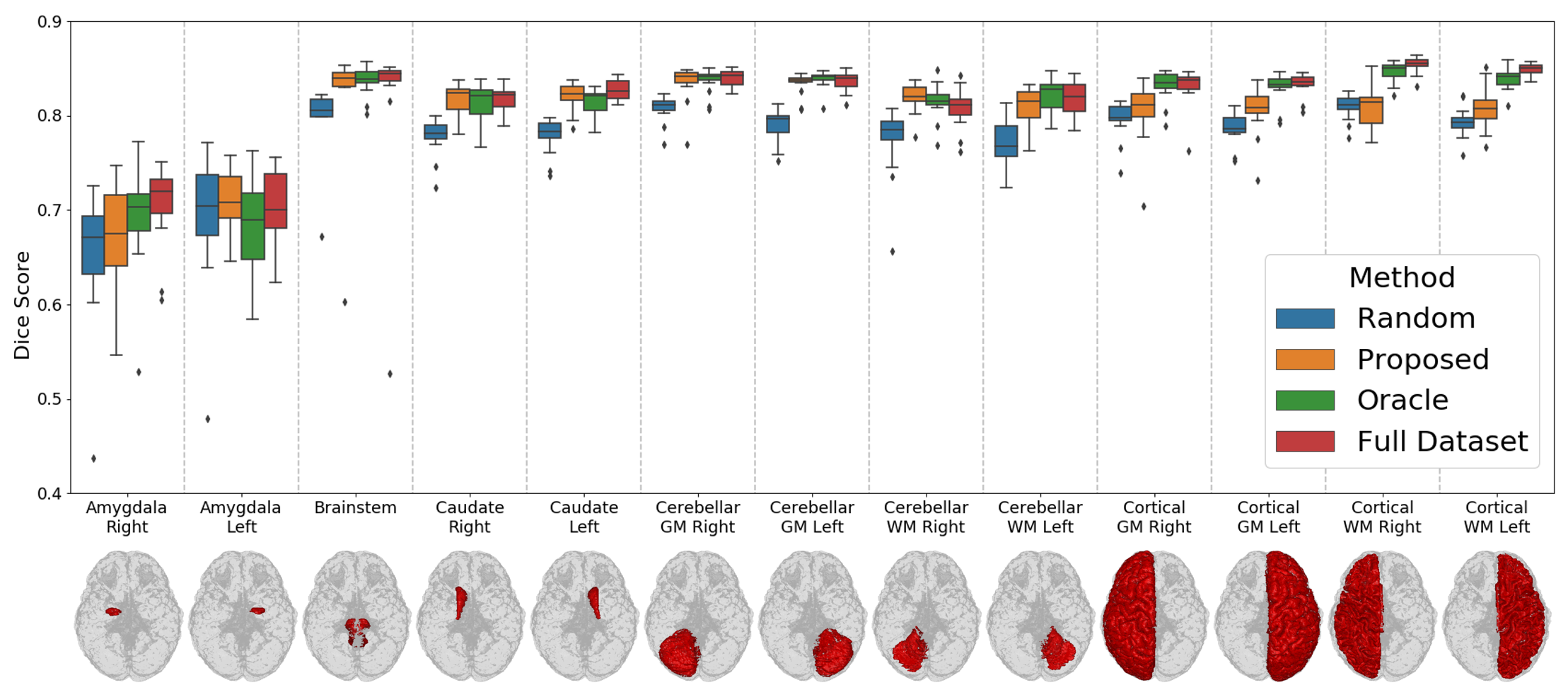}
\includegraphics[width=0.99\textwidth]{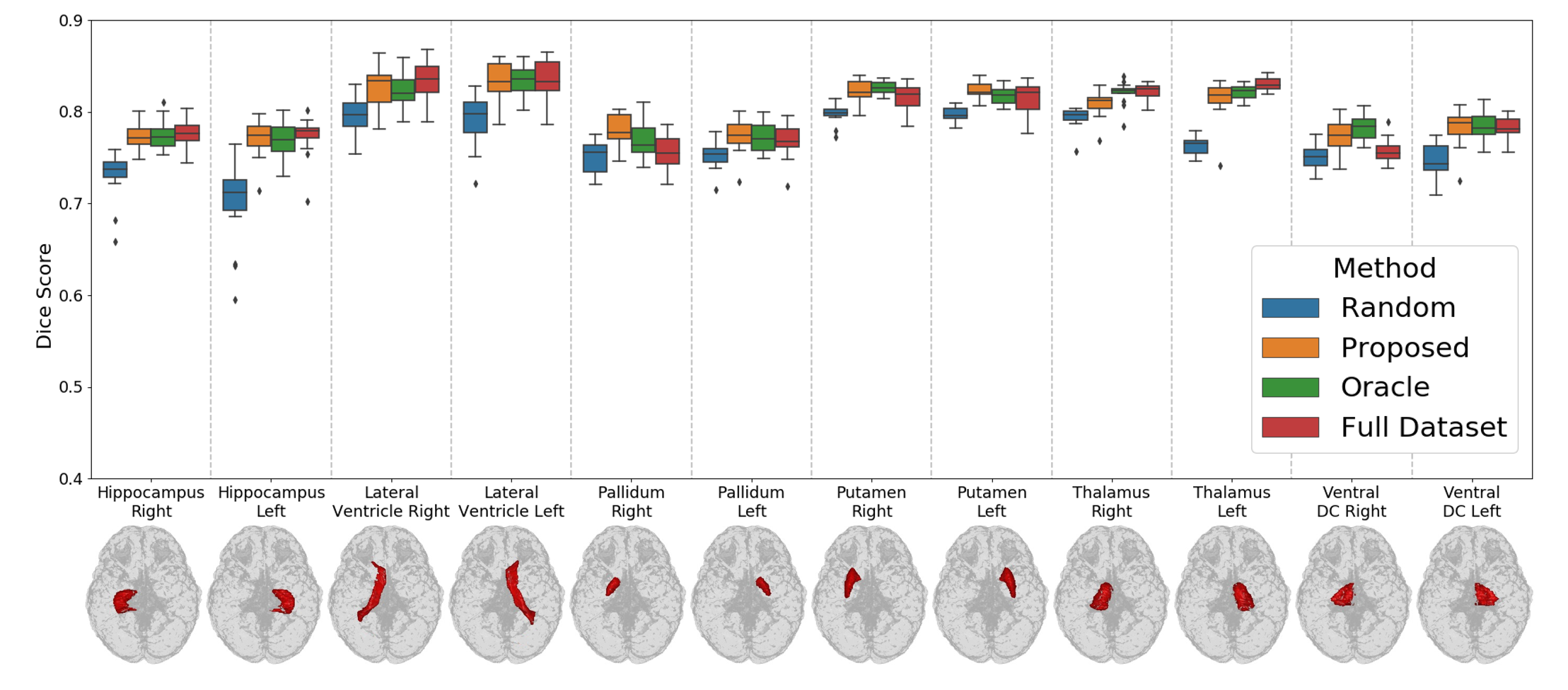}
\caption{Boxplots of the Dice scores for 25 brain structures on MALC test dataset, comparing the proposed method training on 42\% of training set with the random method, oracle method and training on full dataset.} \label{malcresult_1}
\end{figure*}

We then evaluated our method on the MALC dataset for whole brain segmentation task, in total segmenting 25 brain structures. Fig. \ref{malc_test} compares the performance of the proposed method to the random suggestion method and the oracle method, in terms of the averaged Dice score of 25 brain structures. The performance of all methods improves with the increase of the size of the training set. The proposed method constantly outperforms the random method during the experiment and achieves comparable performance to the oracle method when using 60\% of the training set. When training with 12\% of full MALC dataset, random suggestion yields superior performance to both proposed and oracle method. Similar to the good performance of the random method we observed when comparing with the ClsMC method for suggesting brain tumour images, one plausible reason for this was that less informative samples are more important at the early stage of the model training. Fig. \ref{malcresult_1} compares the performance of proposed method to the baseline, oracle methods for the Dice score of each individual brain structures when using 42\% of training set. The result of training on the full training set of MALC dataset is also shown on Fig. \ref{malcresult_1}. The proposed method outperforms the random method across all brain structures, achieving comparable performance to the oracle method and even close to the segmentation model trained on the full training set.

\begin{figure}
\centering
\includegraphics[width=0.48\textwidth]{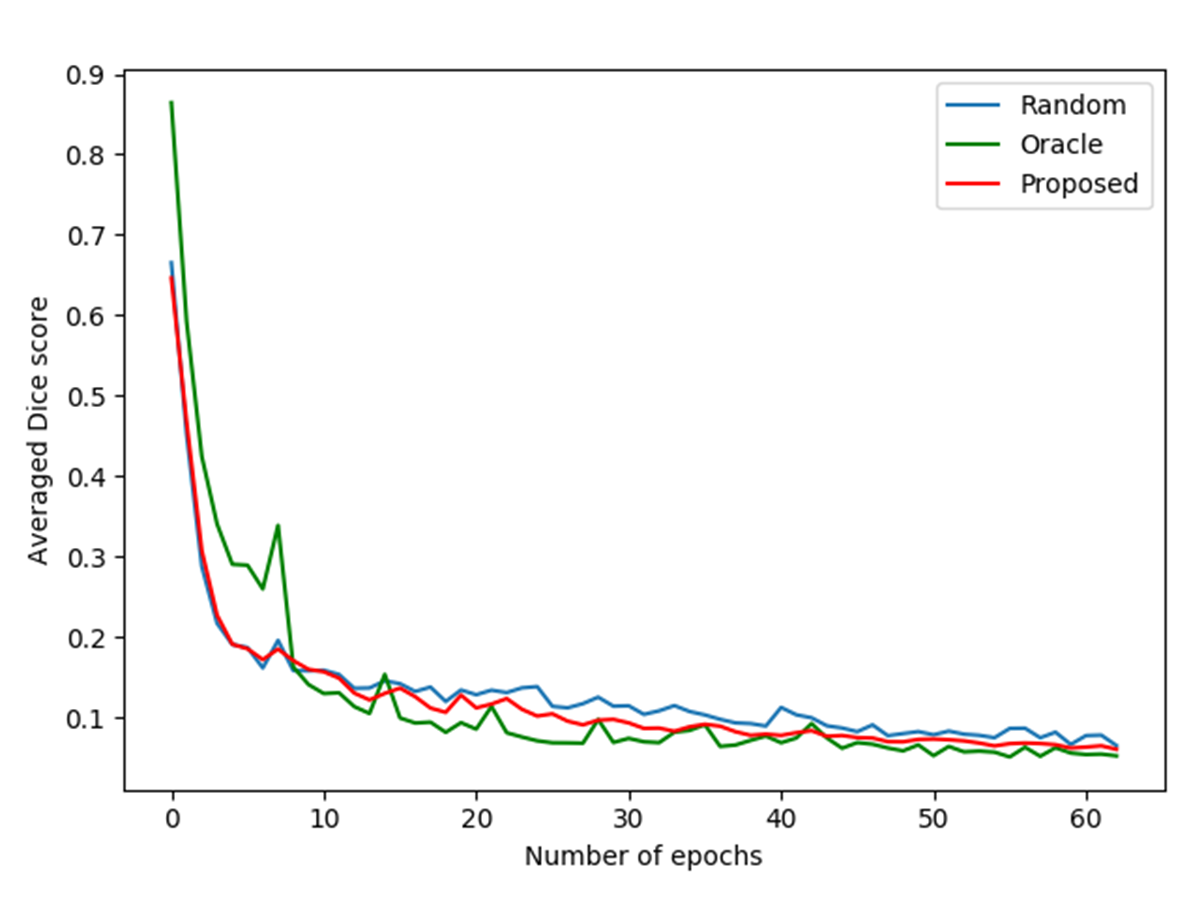}
\caption{Comparison of training loss curves for the proposed suggestive annotation method with the random suggestion method and the oracle method.} \label{trainingloss}
\end{figure}

The training loss curves of the proposed method, baseline method and oracle method are illustrated in Fig. \ref{trainingloss}. As shown in the figure, in the beginning of the experiment, the training losses of the proposed method and the baseline method decrease at a similar rate. However, with more suggested images being added into the training set, the training loss curve of the proposed method decreases further compared to the baseline method and becomes closer to the oracle method.

Table \ref{tab:malc_dice} compares the performance of the proposed method and other competitive method when training using 24\% and 42\% of MALC dataset.  We also performed the Wilcoxon signed-rank test between two top performing methods among random, DAL, ClsMC, LL and ours for Dice score and Hausdorff distance. It shows that the proposed method achieves better performance than random, DAL, ClsMC and LL methods in terms of the mean Dice score of 25 brain structures for using 42\% of MALC dataset. In terms of the Hausdorff distance, the ClsMC method achieves the best performance but not statistically significant.

\begin{table*}
\centering
\caption{Comparison of the proposed method with other methods in terms of the averaged Dice score, the averaged Hausdorff distance (unit: mm) and standard deviations, using 24\% and 42\% of full dataset. The figures in bold are the best results among random, DAL, ClsMC, LL and our method. The results of training on the full training set and the Wilcoxon signed-rank test between two top performing methods among random, DAL, ClsMC, LL and ours are also given in this table.\label{tab:malc_dice}}
\scalebox{1}{
\begin{tabular}{ccccc}
\hline
\multirow{2}{*}{Methods} & \multicolumn{2}{c}{24\% dataset}       & \multicolumn{2}{c}{42\% dataset}        \\ \cline{2-5} 
                         & Dice score        & Hausdorff distance & Dice score         & Hausdorff distance \\ \hline
Random                   & 0.7427$\pm$0.0425 & 14.21$\pm$2.75     & 0.7679$\pm$0.0231  & 12.15$\pm$2.11     \\
DAL \cite[]{yang2017suggestive}                      & 0.7478$\pm$0.0262 & 12.9$\pm$2.24     & 0.7701$\pm$0.0103  & 10.85$\pm$1.65     \\
ClsMC \cite[]{zheng2019biomedical}                    & 0.7503$\pm$0.0192 & \textbf{11.83$\pm$2.15}    & 0.7713$\pm$0.0127  & \textbf{10.07$\pm$1.59}    \\
LL \cite[]{yoo2019learning}                       & 0.7485$\pm$0.0233 & 13.41$\pm$1.99     & 0.7692$\pm$0.0148  & 11.3$\pm$1.8       \\
Proposed                 & \textbf{0.7519$\pm$0.014} & 12.56$\pm$2.77     & \textbf{0.7782$\pm$0.0098} & 10.21$\pm$1.92     \\
Oracle                   & 0.7742$\pm$0.0076 & 11.69$\pm$2.08     & 0.7802$\pm$0.0064  & 10.39$\pm$1.78     \\
Full dataset             & -                 & -                  & 0.7841$\pm$0.0071  & 9.8$\pm$1.46       \\
p-value                 &$\approx$0.38                   &$\approx$0.05                    & \textless0.01               &$\approx$0.4                \\ \hline
\end{tabular}}
\end{table*}

\section{Discussion}
The proposed method achieved close segmentation results as training on the full dataset. For BraTS dataset, training on 7\% of total image slices suggested from 335 MRI volumes by the proposed method achieved very good result. The significant reduction in the size of the training set is because the MRI volumes often contain similar adjacent image slices, which leaves us a large space for reducing the data redundancy. For a smaller MALC dataset, 42\% of the of total image slices were suggested to achieve a good segmentation performance. Overall, the experimental results suggest that the proposed framework has a strong potential in reducing the annotation cost for medical imaging data, which will be beneficial to the medical AI community, especially to machine learning models \cite[]{razzak2018deep,van2021deep} that rely on training on a large amount of annotated data.

Compared to most of other AL frameworks \cite[]{yang2017suggestive,yoo2019learning,liu2020deepactive,shen2020deep,aghdam2019active} which need to iterate the full unannotated dataset for each time of the suggestion to sample most uncertain and/or most representative ones, the proposed method directly suggests the informative samples by using the gradient of loss acquired from the segmentation model. Although the proposed method requires training a VAE, the learnt data manifold can be re-used for different downstream tasks. Furthermore, selecting uncertain samples and representative samples are usually done in separate steps in many active learning frameworks, but the proposed method is designed to combine these steps and strike a balance between the uncertainty and the representativeness of the selected samples to suggest the samples that are beneficial to model training. A well-learnt data manifold is a good latent representation of the original dataset. The uncertainty is taken into account by leveraging the gradient of loss from the segmentation network.  The experimental results indicate that the proposed method outperforms both random suggestion and other suggestion methods that utilise the training loss \cite[]{yoo2019learning} or the latent representation \cite[]{zheng2019biomedical}. We have successfully shown the proposed method achieves a performance close to the oracle method.

Comparing to other methods that are proposed to reduce the annotation cost and to alleviate the data scarcity challenge, such as semi-supervised learning and data augmentation approaches, our proposed method encourages the idea of human-in-the-loop when training the segmentation network. We believe the experts play an irreplaceable role in providing high quality annotation for medical imaging data and efficiently keeping them involved in the process of training and tuning the segmentation network is beneficial to improving the performance of the segmentation network. On the other hand, experts are susceptible to the quality of the MR images, therefore it may be difficult to implement methods like dataset distillation for MRI segmentation task.

There are a few limitations with the proposed framework and our research. Firstly, the performance of the proposed framework relies on the quality of the data manifold learnt by the VAE. Our experiments suggest that the proposed method is more effective with a data manifold learnt with a relatively low $z$-dimension number. This issue may be alleviated by including more unannoated data samples for training the VAE or by generating multiple gradient-integrated images for the same image in each iteration so that the proposed method can search multiple gradient directions and multiple samples can be suggested based on the previous sample; and a new image that is most frequently suggested based on these gradient-integrated images will be given to the expert for annotation. Secondly, what each dimension of the latent space represents is not explainable, therefore it is difficult to understand the behaviour of the sample suggestion process clearly. This would be an interesting direction to explore in the future. Thirdly, the sampling and segmentation stages in our framework are coupled in the iterations, which means the experts for annotating the suggested images are expected to be involved after each iteration of model training. This requires close collaboration and interaction with experts in the process. Fourthly, some modifications had to be made to the original implementations of DAL, ClsMC and LL methods to adapt our experimental design, it is difficult to guarantee an optimal performance from these methods.

\section{Conclusion}
Here we propose a gradient-guided suggestive annotation method, which can be used to train a segmentation network efficiently with a small number of samples in an active learning manner. Using only 7\% of the BraTS dataset or 42\% of the MALC dataset, it achieves a high segmentation performance that is a comparable to that by training on the full dataset. The proposed framework is generic, and it has the potential to be extended to other medical image analysis tasks to reduce the annotation cost and improve learning efficiency.

\section{Acknowledgement}
This independent research was funded by the NIHR Imperial Biomedical Research Centre (BRC). The views expressed in this publication are those of the author(s) and not necessarily those of the NHS, NIHR or Department of Health. We appreciate the script for 3D renderings of brain structures provided by Prof. Bennett Landmann and Dr. Yuankai Huo \cite[]{slant}.

%%Harvard
\bibliographystyle{model2-names.bst}\biboptions{authoryear}
\bibliography{refs}

\end{document}